\documentclass[10pt,twocolumn,letterpaper]{article}

\usepackage{cvpr}      
\usepackage{url}
\usepackage[dvipsnames]{xcolor}

\definecolor{cvprblue}{rgb}{0.21,0.49,0.74}
\usepackage[pagebackref,breaklinks,colorlinks,citecolor=cvprblue]{hyperref}

\def\paperID{*****} 
\def\confName{CVPR}
\def\confYear{2024}

\title{DreaMoving: A Human Video Generation Framework \\based on Diffusion Models}

\author{Mengyang Feng, Jinlin Liu, Kai Yu, Yuan Yao, Zheng Hui, Xiefan Guo, \\ Xianhui Lin, Haolan Xue, Chen Shi, Xiaowen Li, Aojie Li, Xiaoyang Kang, Biwen Lei, \\Miaomiao Cui, Peiran Ren, Xuansong Xie\\
Alibaba Group\\
{\tt\small \{mengyang.fmy, ljl191782, jinmao.yk, ryan.yy, huizheng.hz, guoxiefan.gxf, } \\ {\tt\small xianhui.lxh, haolan.xhl, zhicheng.sc, lxw262398, liaojie.laj, kangxiaoyang.kxy, biwen.lbw} \\
 {\tt\small miaomiao.cmm, peiran.rpr, xingtong.xxs\}@alibaba-inc.org}
}

\begin{document}

\maketitle

\begin{abstract}
In this paper, we present DreaMoving, a diffusion-based controllable video generation framework to produce high-quality customized human videos. Specifically, given target identity and posture sequences, DreaMoving can generate a video of the target identity moving or dancing anywhere driven by the posture sequences. To this end, we propose a Video ControlNet for motion-controlling and a Content Guider for identity preserving. The proposed model is easy to use and can be adapted to most stylized diffusion models to generate diverse results. The project page is available at \url{https://dreamoving.github.io/dreamoving}.

\end{abstract}

\section{Introduction}  
\label{sec:related-work}
Recent text-to-video (T2V) models like Stable-Video-Diffusion\footnote{\url{https://stability.ai/news/stable-video-diffusion-open-ai-video-model}} and Gen2\footnote{\url{https://research.runwayml.com/gen2}} make breakthrough progress in video generation. However, it is still a challenge for human-centered content generation, especially character dance. The problem involves the lack of open-source human dance video datasets and the difficulty of obtaining the corresponding precise text description, making it a challenge to train a T2V model to generate videos with intraframe consistency, longer length, and diversity. Besides, personalization and controllability stand as paramount challenges in the realm of human-centric content generation, attracting substantial scholarly attention. Representative research like ControlNet~\cite{zhang2023adding} is proposed to control the structure in the conditional image generation, while DreamBooth~\cite{ruiz2022dreambooth} and LoRA~\cite{hu2022lora} show the ability in appearance control through images. However, these techniques often fail to offer precise control over motion patterns or necessitate hyperparameter fine-tuning specific to target identities, introducing an additional computational burden. Customized video generation is still under investigation and represents a largely uncharted territory. 
In this paper, we present a human dance video generation framework based on diffusion models (DM), named \textbf{DreaMoving}.

The rest of the paper is organized as follows. Sec.~\ref{sec:arch} presents a detailed description of how the DreaMoving is built. Sec.~\ref{sec:res} presents some results generated by our method.

%

%

\section{Architecture}
\label{sec:arch}
\begin{figure*}
  \centering
  \includegraphics[width=0.95\linewidth]{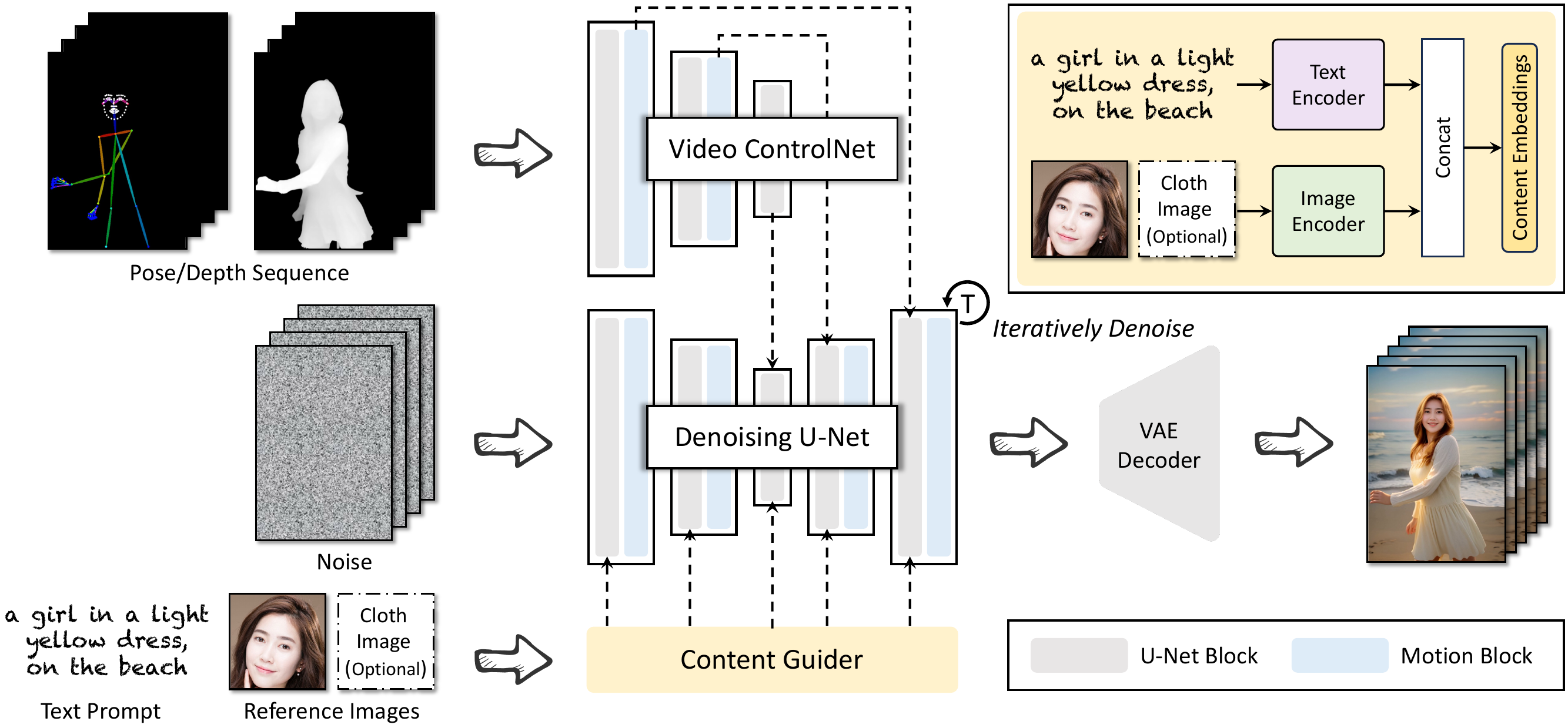}
  \caption{The overview of DreaMoving. The Video ControlNet is the image ControlNet~\cite{zhang2023adding} injected with motion blocks after each U-Net block. The Video ControlNet processes the control sequence (pose or depth) to additional temporal residuals. The Denoising U-Net is a derived Stable-Diffusion~\cite{rombach2021highresolution} U-Net with motion blocks for video generation. The Content Guider transfers the input text prompts and appearance expressions, such as the human face (the cloth is optional), to content embeddings for cross attention.   
}
  \label{fig:pipeline}
\end{figure*}

DreaMoving is built upon Stable-Diffusion~\cite{rombach2021highresolution} models. As illustrated in Fig.~\ref{fig:pipeline}, it consists of three main networks, including the Denoising U-Net, the Video ControlNet, and the Content Guider. Inspired by AnimateDiff~\cite{guo2023animatediff}, we insert motion blocks after each U-Net block in the Denoising U-Net and the ControlNet. The Video ControlNet and the Content Guider work as two plug-ins of the Denoising U-Net for controllable video generation.
The former is responsible for motion-controlling while the latter is in charge of the appearance representation.

\subsection{Data Collection and Preprocessing}
To gain better performance in generating human videos, we collected around 1,000 high-quality videos of human dance from the Internet. As the training of the temporal module needs continuous frames without any transitions and special effects, we further split the video into clips and finally got around 6,000 short videos (8$\sim$10 seconds). For text description, we take Minigpt-v2~\cite{chen2023minigpt} as the video captioner. Specifically, using the ``grounding'' version, the instruction is \emph{[grounding] describe this frame in a detailed manner}. The generated caption of the centered frame in keyframes represents the whole video clip's description, mainly describing the content of the subject and background faithfully.

\subsection{Motion Block}
To improve the temporal consistency and motion fidelity, we integrate motion blocks into both the Denosing U-Net and ControlNet. The motion block is extended from the AnimateDiff~\cite{guo2023animatediff}, and we enlarge the temporal sequence length to 64. We first initialize the weights of motion blocks from the AnimateDiff (\emph{mm\underline{~~}sd\underline{~~}v15.ckpt}) and fine-tune them on the private human dance video data.

\subsection{Content Guider}
The Content Guider is designed to control the content of the generated video, including the appearance of human and the background. One simple way is to describe the human appearance and background with a text prompt, such as 'a girl in a white dress, on the beach'. However, it is hard to describe a personalized human appearance for a normal user. Even by complex prompt engineering, the model may not give the desired output.

Inspired by IP-Adapter~\cite{ye2023ip-adapter}, we propose to utilize the image prompt for precise human appearance guidance and the text prompt for background generation. Specifically, a face image is used to encode the facial features through an image encoder, and a cloth/body image is optionally involved to encode the body features. The text and human appearance features are concatenated as the final content embeddings. The content embeddings are then sent to cross-attention layers for human appearance and background representations as described in IP-Adapter~\cite{ye2023ip-adapter}.
Given the query features $Z$, the text features $c_t$, the face features $c_f$, and the cloth features $c_c$, the output of cross-attention $Z^{'}$ can be defined by the following equation:

\begin{equation}
\label{eqn:Content-Guider}
    \begin{split}
        Z' &= soft\max \left( {\frac{{QK_t^T}}{{\sqrt d }}} \right){V_t} + {\alpha _f}soft\max \left( {\frac{{QK_f^T}}{{\sqrt d }}} \right){V_f} \\&+ {\alpha _c}soft\max \left( {\frac{{QK_c^T}}{{\sqrt d }}} \right){V_c},
    \end{split}
\end{equation}
where, $Q = ZW_q$, $K_t = c_t W^t_k$, and $V_t = c_t W^t_v$  are the query, key, and values matrices from the text features, $K_f = c_f W^f_k$, and $V_f = c_f W^f_v$  are the key, and values matrices from the face features, and $K_c = c_c W^c_k$, and $V_c = c_c W^c_v$  are the key, and values matrices from the cloth features. $\alpha _f$ and $\alpha _c$ are the weights factor.

\subsection{Model Training}

\subsubsection{Content Guider Training}
The Content Guider serves as an independent module for base diffusion models. Once trained, it can generalized to other customized diffusion models. We trained Content Guider based on SD v1.5 and used OpenCLIP ViT-H14~\cite{ilharco_gabriel_2021_5143773} as the image encoder as \cite{ye2023ip-adapter}. To better preserve the identity of the reference face, we employ the Arcface~\cite{deng2019arcface} model to extract the facial correlated features as a supplement to the clip features. We collect the human data from LAION-2B, then detect and filter images without faces. During training, the data are randomly cropped and resized to $512\times512$. Content Guider is trained on a single machine with 8 V100 GPUs for 100k steps, batch size is set to 16 for each GPU, AdamW optimizer~\cite{loshchilov2017decoupled} is used with a fixed learning rate of $1e-4$ and weight decay of $1e-2$. 

\subsubsection{Long-Frame Pretraining}
We first conduct a warming-up training stage to extend the sequence length in the motion module from 16 to 64 on the validation set (5k video clips) of WebVid-10M~\cite{Bain21}. We only train the motion module of the Denoising U-Net and freeze the weights of the rest of the network. No ControlNet and image guidance are involved in this stage. 
The learning rate is set to $1e-4$ and the resolution is $256\times256$ (resize \& center crop). The training is stopped after 10k steps with a batch size of 1.

\subsubsection{Video ControlNet Training}
After the long-frame pretraining, we train the Video ControlNet with the Denoising U-Net by unfreezing both the motion blocks and U-Net blocks in the Video ControlNet and fixing the Denoising U-Net. The weights of motion blocks in Video ControlNet are initialized from the long-frame pretraining stage. In this stage, we train the network on the collected 6k human dance video data. No image guidance is involved in this stage. The human pose or depth is extracted as the input of the Video ControlNet using DWPose~\cite{yang2023effective} or ZoeDepth~\cite{https://doi.org/10.48550/arxiv.2302.12288}, respectively. 
The learning rate is set to $1e-4$ and the resolution is $352\times352$. The training is stopped after 25k steps with a batch size of 1.

\subsubsection{Expression Fine-tuning}
To gain better performance in human expression generation, we further fine-tune the motion blocks in Denoising U-Net by training with the Video ControlNet on the collected 6k human dance video data. In this stage, the whole Video ControlNet and the U-Net blocks in the Denoising U-Net are locked, and only the weights of the motion blocks in the Denoising U-Net are updated. 
The learning rate is set to $5e-5$ and the resolution is $512\times512$. The training is stopped after 20k steps with a batch size of 1.

\subsection{Model Inference}
At the inference stage, the inputs are composed of the text prompt, the reference images, and the pose or depth sequence. The control scale of the Video ControlNet is set to 1.0 for pose or depth only. Our method also supports the form of multi-controlnet, and the depth and pose Video ControlNets can be used simultaneously. The strength of the face/body guidance is also controllable in the Content Guider by adjusting the $\alpha _f$ and $\alpha _c$ in Eqn.~\ref{eqn:Content-Guider}. The content is fully controlled by the text prompt if $\alpha _f = \alpha _c = 0$.

\section{Results}
\label{sec:res}
\begin{figure*}
  \centering
  \includegraphics[width=1.0\linewidth]{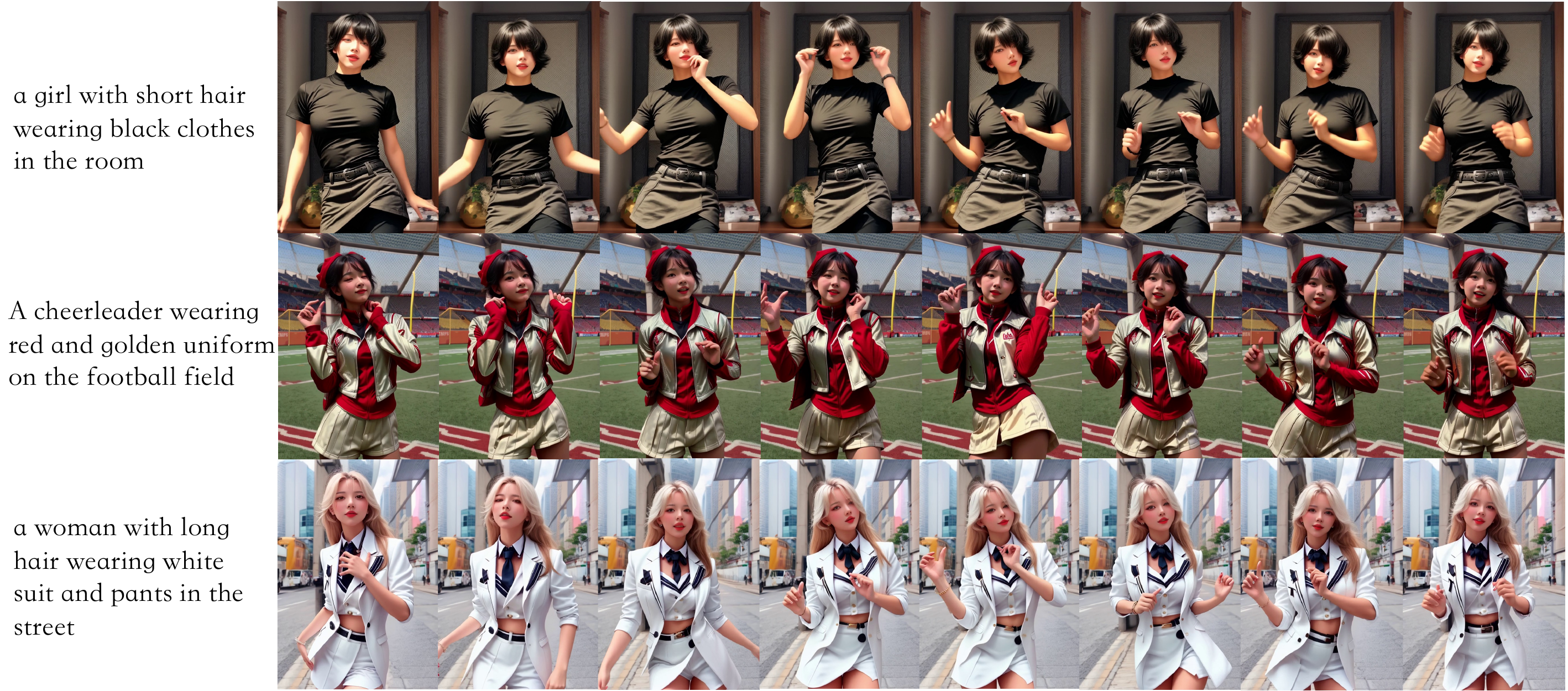}
  \caption{The results of DreaMoving with text prompt as input.}
  \label{fig:text-prompt-only}
\end{figure*}

\begin{figure*}
  \centering
  \includegraphics[width=1.0\linewidth]{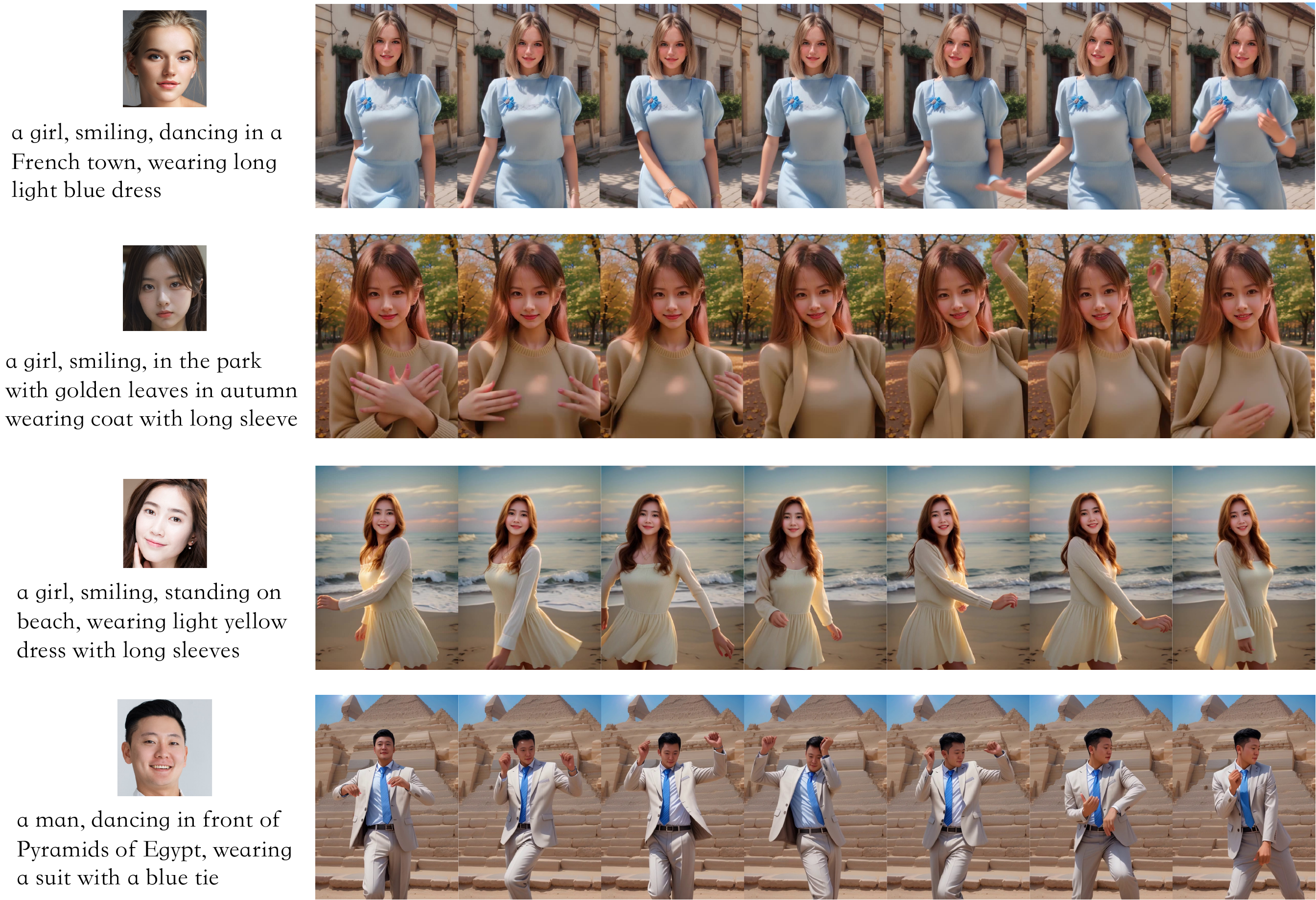}
  \caption{The results of DreaMoving with text prompt and face image as inputs.}
  \label{fig:text-and-face-prompts}
\end{figure*}

\begin{figure*}
  \centering
  \includegraphics[width=1.0\linewidth]{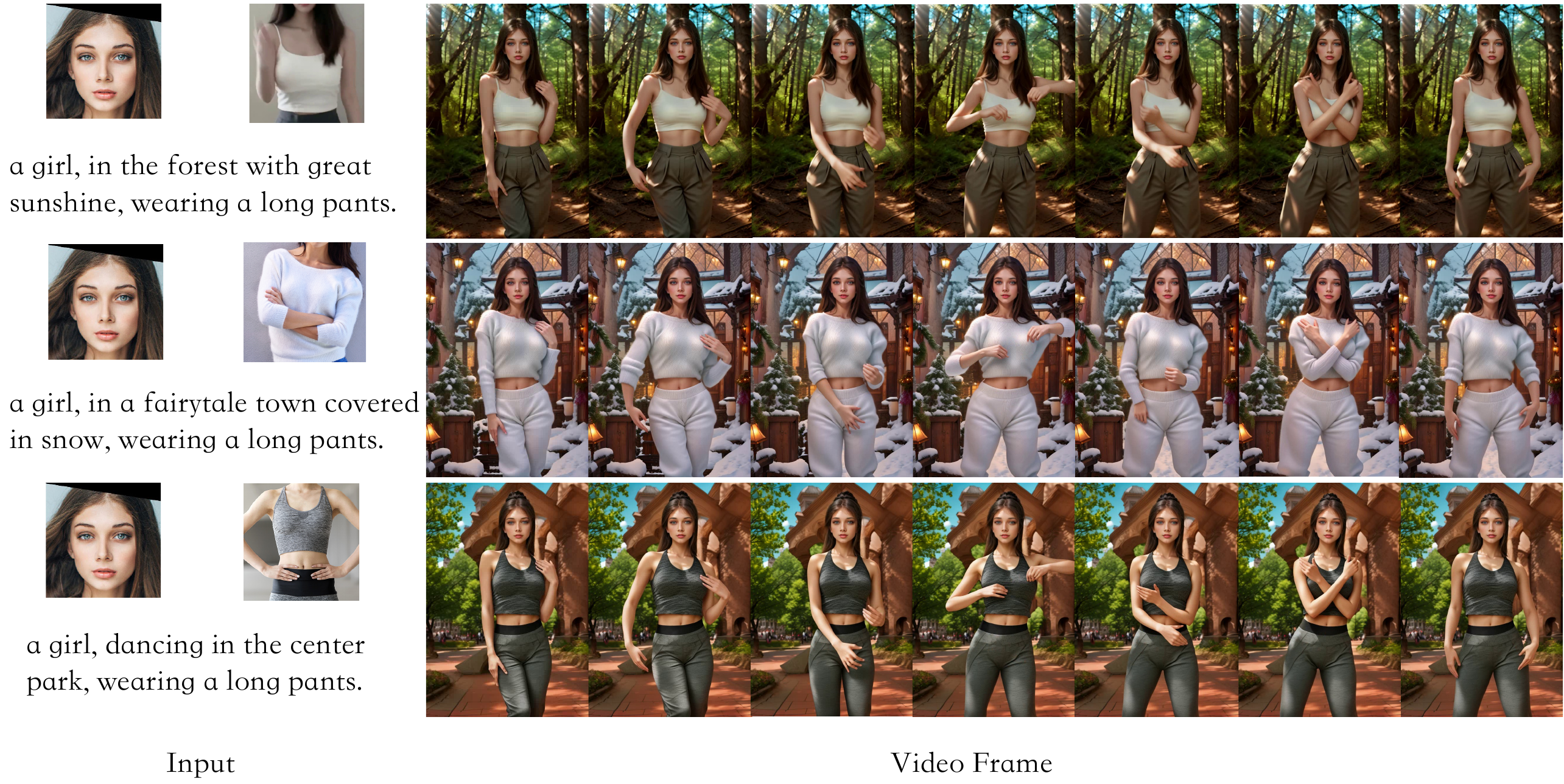}
  \caption{The results of DreaMoving with face and cloth images as inputs.}
  \label{fig:text-and-face-and-cloth-prompts}
\end{figure*}

\begin{figure*}
  \centering
  \includegraphics[width=0.9\linewidth]{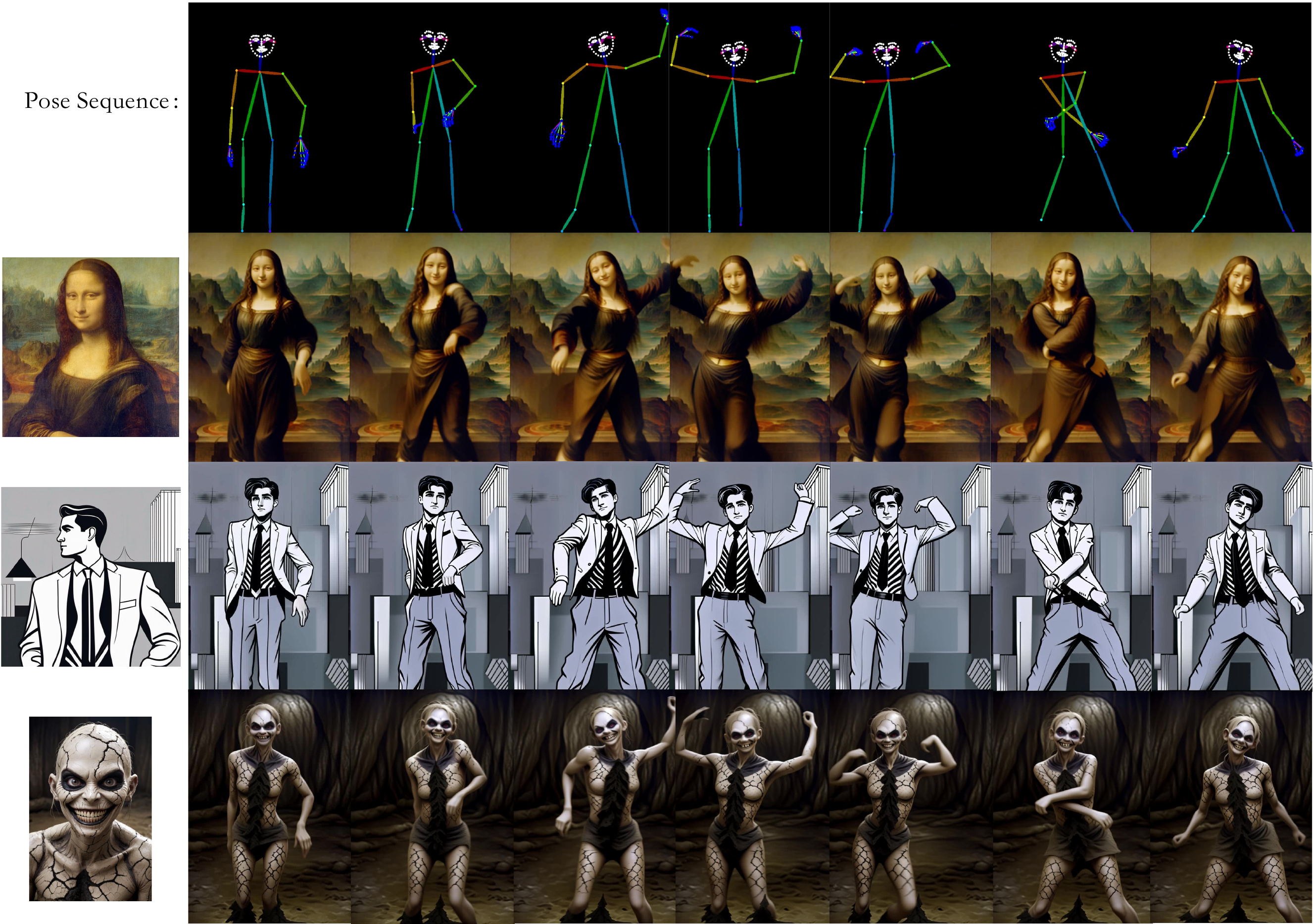}
  \caption{The results of DreaMoving with stylized image as input.}
  \label{fig:image-prompt-only}
\end{figure*}

DreaMoving can generate high-quality and fidelity videos given guidance sequence and simple content description (text prompt only, image prompt only, or text-and-image prompts) as input. In Fig.~\ref{fig:text-prompt-only}, we show the result with text prompt only. To keep the face identity, the user can input the face image to the Content Guider to generate a video of some specific person (demonstrated in Fig.~\ref{fig:text-and-face-prompts}). Moreover, the user can define both the face content and clothes content, as exhibited in Fig.~\ref{fig:text-and-face-and-cloth-prompts}. We further test the generalization of the proposed method on images of unseen domains. In Fig.~\ref{fig:image-prompt-only}, we run DreaMoving using unseen stylized images. Our method is able to generate videos in accordance with the style and content in the input image. 


{
    \small
    \bibliographystyle{ieeenat_fullname}
    \bibliography{main}
}


\end{document}